\documentclass[letterpaper, 10 pt, conference]{ieeeconf}  

\IEEEoverridecommandlockouts                              
\usepackage{graphics} 
\usepackage{epsfig} 
\usepackage{mathptmx} 
\usepackage{times} 
\usepackage{amsmath} 
\usepackage{amssymb}  
\usepackage[dvipsnames]{xcolor} 
\usepackage{multicol}
\usepackage{glossaries}
\usepackage{siunitx}
\usepackage[hidelinks]{hyperref}
\usepackage[super]{nth}
\usepackage{bm}
\usepackage{physics}

\usepackage[dvipsnames]{xcolor}

\DeclareMathAlphabet{\mathcal}{OMS}{cmsy}{m}{n}

\title{\LARGE \bf Learning Low-Frequency Motion Control for\\Robust and Dynamic Robot Locomotion}

\author{
    Siddhant Gangapurwala, Luigi Campanaro and Ioannis Havoutis
    \thanks{The authors are with Dynamic Robots Systems
    (DRS) group, Oxford Robotics Institute, University
    of Oxford, UK. Email: {\tt\footnotesize \{siddhant,
    luigi, ioannis\}@robots.ox.ac.uk}}
}

\begin{document}

\maketitle
\thispagestyle{empty}
\pagestyle{empty}

\begin{abstract}
    Robotic locomotion is often approached with the goal of maximizing
    robustness and reactivity by increasing motion control frequency. We challenge this intuitive notion
    by demonstrating 
    robust and dynamic locomotion with a learned motion controller executing at as
    low as \SI{8}{\hertz} on a real ANYmal C quadruped. The robot is able
    to robustly and repeatably achieve a high heading velocity of \SI{1.5}{\meter\per\second}, 
    traverse uneven terrain, and resist unexpected external perturbations.
    We further present a comparative analysis of deep reinforcement learning (RL) based
    motion control policies trained and executed at frequencies ranging from \SI{5}{\hertz}
    to \SI{200}{\hertz}. We show that low-frequency policies are less sensitive
    to actuation latencies and variations in system dynamics. 
    This is to the extent that a successful sim-to-real transfer can 
    be performed even without any dynamics randomization or actuation modeling.
    We support this claim through a set of rigorous empirical evaluations. Moreover, to assist reproducibility, we provide
    the training and deployment code along with an extended analysis at \url{https://articulated.robots.ox.ac.uk/lfmc/}.
\end{abstract}

\section{Introduction}
\label{sec:introduction}
Legged systems can execute agile motions by leveraging their ability to reach
appropriate and disjoint support contacts, thereby enabling outstanding mobility in complex and unstructured environments.
This, however, requires control solutions that are able to
recover from unexpected perturbations, adapt to variations in system and environment dynamics,
and execute safe and reliable locomotion. For feedback-based control
systems, taking a corrective control action as soon as a sensory
signal is detected allows for minimizing motion tracking errors while offering high reactivity
to address external disturbances and modeling inaccuracies. This design motivation has been
employed for achieving dynamic locomotion 
behaviors in~\cite{di2018dynamic, Bellicoso2018, yang2022fast}
through generation of low-level actuation commands at 
frequencies ranging from \SI{400}{\hertz} to \SI{1}{\kilo\hertz}.

In contrast, animals are able to demonstrate remarkably agile locomotion in spite
of sensory noise~\cite{faisal2008noise} and considerable sensorimotor latencies~\cite{more2010scaling}
associated with nerve conduction, 
electro-mechanical, and force generation delays~\cite{more2013sensorimotor} which limits
their motion control frequency.
The sensing and actuation delays for a medium-sized 
\SI{20}{\kilo\gram} dog, for example, can be 
approximately \SI{58}{\milli\second} of which
\SI{23.2}{\milli\second} are required to process sensory feedback, 
generate an actuation signal, and deliver electro-mechanical commands~\cite{more2018scaling}. 
The remaining delay 
corresponds to the ramp-up time for achieving peak muscle force. 
For a \SI{40}{\kilo\gram} animal, the total delay is estimated to be \SI{67}{\milli\second}
with processing, generation and delivery latency of \SI{30.4}{\milli\second}.

In~\cite{ashtiani2021hybrid}, \textit{Ashtiani et al.} present an example in which
a house cat exhibiting a locomotion frequency of \SI{5}{\hertz}~\cite{bertram2014domestic}
is sensor-blind for half its stance-phase. This duration corresponds to
the entirety of the muscle force ramp-up time suggesting that 
high-frequency feedback-based decision-making may not be critical 
for locomotion over challenging terrains. \textit{Ashtiani et al.} 
investigate this discrepancy between biological and mechanical
systems and propose a parallel compliant joint system along
with a leg-length controller to realize actuation response similar to 
that of animal muscle-tendon units. This is based on the motivation that 
elastic actuation allows for self-stability~\cite{blickhan2007intelligence}.

In this regard, the ANYmal C quadruped~\cite{anymalc}
houses series elastic actuators (SEAs) that offer high compliance making
the system robust to impacts. SEAs, however, trade-off controllability
for compliance~\cite{pratt1995series}.
In comparison, quasi-direct drives offer better actuation command tracking performance with
lower control latencies enabling highly dynamic locomotion~\cite{bledt2018cheetah}. This makes
it possible for the Mini Cheetah to sprint at \SI{3.74}{\meter\per\second}~\cite{ji2022concurrent}
while for the ANYmal C, \textit{Miki et al.} reported heading and lateral 
velocities of up to \SI{1.2}{\meter\per\second} even with an extremely
robust locomotion controller~\cite{miki2022learning}.

\begin{figure}
    \centering
    \includegraphics[width=0.45\textwidth]{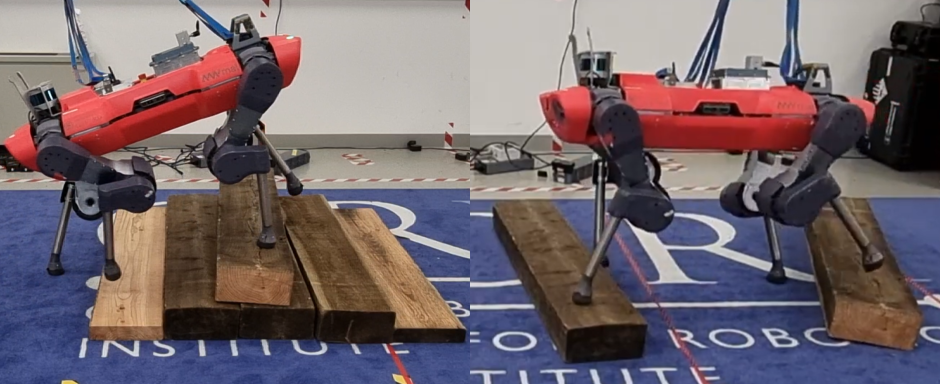}
    \caption{ANYmal C quadruped walking over uneven terrain with a perceptive reinforcement learning policy executed at a 
        frequency of \SI{8}{\hertz}. Accompanying video can be found at \url{https://youtu.be/pSuX223zLvM}}
    \label{fig:anymal_c_lab_terrain}
    \vspace{-0.2cm}
\end{figure}

This work explores and presents our findings, alluding to a bio-inspired control 
design choice: \textit{if animals
can perform robust and dynamic locomotion at low motion control frequencies, can
robots do so too?} 

For this, we develop blind and perceptive control strategies 
for the ANYmal C quadruped and evaluate its performance for
robust and dynamic locomotion over flat and uneven terrain 
as shown in Fig.~\ref{fig:anymal_c_lab_terrain}.

\subsection{Related Work}
Model-free data-driven deep reinforcement learning (RL) enables 
obtaining control solutions that have the potential to 
thoroughly utilize the system capabilities of current robots. This property
has been leveraged for learning 
agile and dynamic robotic 
locomotion skills to perform blind bipedal traversal
over stairs~\cite{siekmann2021blind}, quadrupedal locomotion over challenging
terrains~\cite{Leeeabc5986} and even robust quadrupedal state
recovery~\cite{yang2020multi}. Model-based techniques have also demonstrated dynamic and complex 
locomotion~\cite{grandia2019feedback, jenelten2020perceptive, kim2020vision}.
A combination of model-based and model-free methods have also 
been proposed~\cite{zhang2016learning, Xie2018, gangapurwala2020rloc, gangapurwala2021real}.
These approaches, however, often employ finite-order motion parameterization which inhibits
the discovery of optimal behaviors. In contrast, motions
executed by RL policies that map robot state information to desired
joint states are not constrained by motion primitives. This
makes RL particularly suitable for our task 
of obtaining motion control policies operating at 
frequencies as low as \SI{5}{\hertz}. In such a case,
the optimal behavior is not bounded by carefully tuned
model-based controllers. Instead, the objective of finding
the appropriate style to achieve dynamic and stable
locomotion is addressed by the RL agent.

Despite the significant progress in RL for robotic 
locomotion, there remains an
inconsistency in the design motivations for
much of the proposed control architectures.
In~\cite{hwangbo2019learning}, \textit{Hwangbo et al.} train an
RL locomotion policy to map robot states to desired joint
positions. This policy is queried at \SI{200}{\hertz} and the authors
especially note that introducing a history of joint states into the RL state space
is essential to obtain a locomotion policy.
\textit{Rudin et al.}, however, train a
locomotion policy at \SI{50}{\hertz} without
utilizing joint state history~\cite{rudin2022learning}. The obtained
policy is transferable to the real platform even with 
access to only the current proprioceptive state. 
In~\cite{duan2021learning}, \textit{Duan et al.} also show
bipedal locomotion at \SI{40}{\hertz}
without utilizing joint state history.
We study these differences and observe that at higher motion control frequencies,
the controller is more sensitive to the actuation dynamics compared to
at lower-frequencies. In the context of this
work, low-frequency refers to \SI{25}{\hertz} or less. For a swing and stance phase duration of 
approximately \SI{600}{\milli\second} during locomotion, this corresponds to 
15 or less control set points generated during each of these phases. 
We also observed that at less than \SI{5}{\hertz}, corresponding to less 
than 3 set points generated during swing phase, the controller is unable to 
maintain stability during locomotion. We
detail upon our findings in Section~\ref{sec:results-and-discussion} of
this manuscript.

It is also worth mentioning that a rich body of work focuses on bio-inspired 
mechanical designs~\cite{saputra2021aquro, ijspeert2008central, coyle2018bio}. 
Although this is beyond the scope of our current work, we believe it serves
as an important reminder that control intelligence and mechanical design
complement each other~\cite{pfeifer2012challenges}.

\subsection{Contribution}
Our main contribution with this work is presenting
that low-frequency motion control is sufficient to perform robust
and dynamic locomotion. We further show that
dynamics randomization or actuation modeling 
may not even be necessary for successful sim-to-real
transfer. We additionally
provide a comparative analysis of motion control policies trained
and deployed at a range of frequencies. We believe this work will provide an
important reference to the robotic control
research community with regards to design
motivations for developing custom control solutions.

We additionally highlight our contributions
relating to sharing of training and deployment
code, the low-frequency motion control (LFMC) framework. We
provide a RaiSim~\cite{raisim} based optimized implementation 
for training locomotion policies for ANYmal C at various motion
control frequencies. This allows users to train policies in 
less than thirty minutes on a standard computer without requiring expensive hardware for massive parallelization~\cite{rudin2022learning}.
We additionally provide minimal deployment code
in both C++ and Python. For the Python version, we also provide an
option to choose between the RaiSim and PyBullet~\cite{coumans2021} simulation
engines. We hope this
encourages reproducibility and allows colleagues 
to easily perform benchmarking against our approach.





\section{Preliminaries}
\label{sec:preliminaries}
\subsection{System Model}
    We model a quadrupedal robot as a floating base $B$ with four attached limbs. The robot state
    is measured and expressed in a global reference frame where the position is written as
    $\mathbf{r}_{B}\in\mathbb{R}^3$. The orientation is represented as the 
    rotation matrix $\mathbf{R}_{B}\in\mathit{SO}(3)$. Each limb comprises three rotational joints.
    The angular joint positions are denoted by the vector $\mathbf{q}_{j}\in\mathbb{R}^{12}$.
    The linear and angular base velocities are represented as
    $\mathbf{v}_{B}\in\mathbb{R}^3$ and $\mathbf{\omega}_{B}\in\mathbb{R}^3$
    respectively.
    
    The joint control torques $\boldsymbol{\tau}_j\in\mathbb{R}^{12}$ actuate the quadrupedal system and are
    computed using the impedance control model,
    
    \begin{equation}
        \boldsymbol{\tau}_j = k_p\Delta\mathbf{q}_j - k_d\mathbf{\dot{q}}_j.
    \label{eq:impedance_controller_simplified}
    \end{equation}
    Here, $k_p$ and $k_d$ represent tracking gains and $\Delta\mathbf{q}_j=\mathbf{q}^{\ast}_j-\mathbf{q}_j$ where
    $\mathbf{q}^{\ast}_j$ denotes the desired joint positions.

\subsection{Control Architecture}
    \begin{figure}
        \centering
        \includegraphics[width=0.48\textwidth]{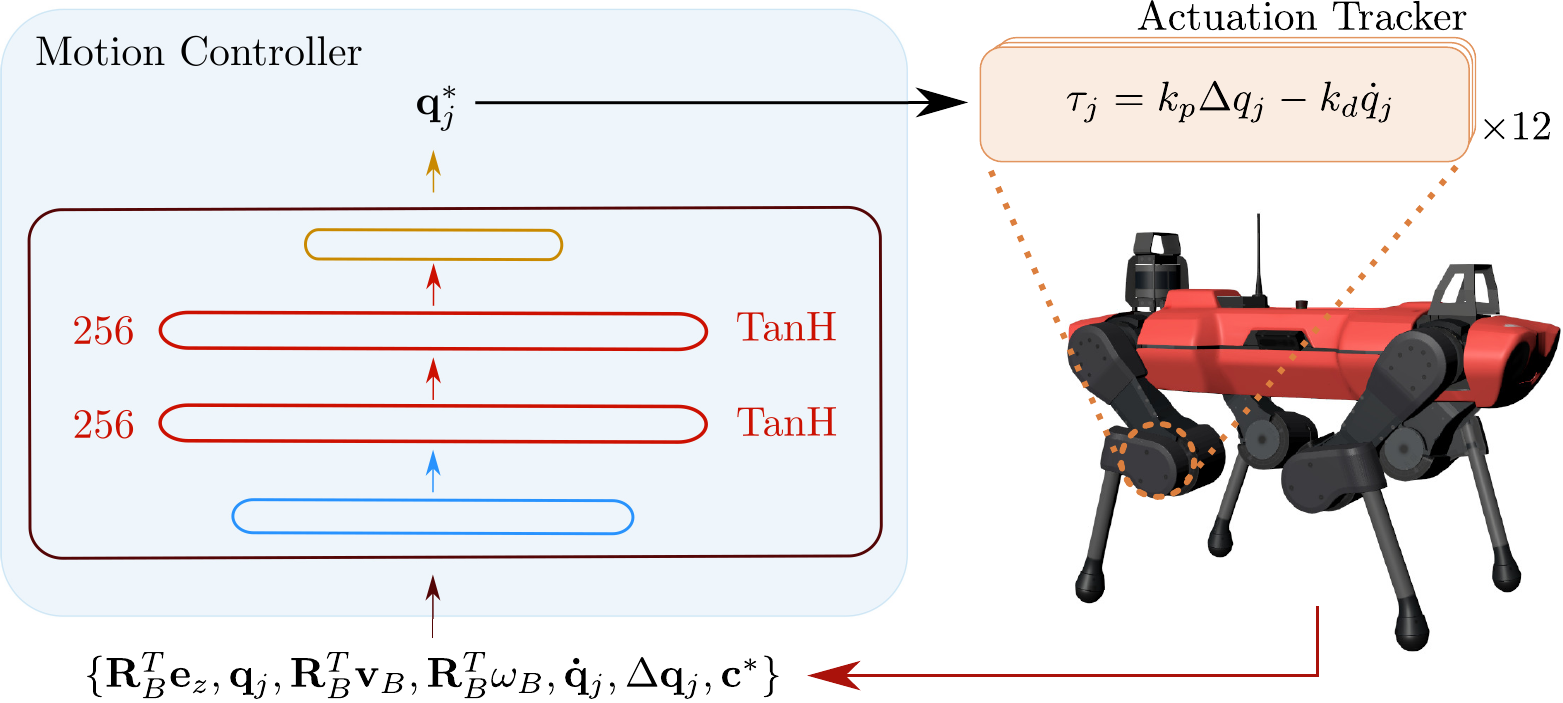}
        \caption{Control architecture of our proprioceptive locomotion framework 
        comprising a motion controller and an
        actuation tracker.}
        \label{fig:control_architecture}
        \vspace{-0.3cm}
    \end{figure}

    Our control architecture comprises a high-level \textit{motion controller} and a low-level \textit{actuation tracker}. 
    This design is motivated by prior works on RL for robotic locomotion~\cite{hwangbo2019learning, gangapurwala2020guided, xie2021dynamics}.
    The motion controller, executed at a deployment frequency $f_{m}$, processes robot state information to
    generate desired joint states. The actuation tracker, executed at a frequency $f_{a}$ where $f_{a}\ge f_{m}$, 
    tracks these desired joint states by generating $\boldsymbol{\tau}_j$ using the model
    described in Eq.~\ref{eq:impedance_controller_simplified}. 

    We model the motion controller policy as a multi-layer perceptron (MLP), $\pi_{\theta}$. Here,
    $\theta$ represents the network parameters. The policy, $\pi_{\theta}: \mathbf{s}\mapsto\mathbf{a}$, 
    maps the input state tuple $\mathbf{s}$ to actions $\mathbf{a}\in\mathbb{R}^{12}$. 
    The tuple $\mathbf{s}$ comprises observations that can be accessed on the real robot. 
    Since we perform comparative analysis of different types of policies, the dimensionality of $\mathbf{s}$
    depends on the individual motion control policy. We discuss this in the following subsection.
    
    Each of the policies outputs an action tuple,
    $\mathbf{a}:=\langle \mathbf{q}^{\ast}_j \rangle$,
    representing the desired
    joint positions and
    is based on the motivation that low-impedance joint position control can offer improved
    training and control performance over 
    torque control~\cite{peng2017learning}.

\subsection{Motion Control Policies}
    We represent the motion control policies as $\pi_{\theta}$ where
    $\theta$ denotes the parameters of a 
    generic motion controller. To refer to specific policies, we introduce the
    notation
    \begin{equation*}
        \pi^{f_t}_{M:H}
    \label{eq:policy_notation}
    \end{equation*}
    where $f_t$ is the motion control frequency at which the policy was trained,
    $M$ is the mode of operation which can either be $b$ (for blind) or $p$
    (for perceptive),
    and $H\in\mathbb{R}$ represents history length of joint states introduced
    in the state tuple $\mathbf{s}$.
    The joint state history is recorded at a frequency of $f_j$ with a 
    corresponding time step $t_j$. 
    In this work, the joint state recording frequency $f_j\ge{}f_m$. We use 
    $f_j=\SI{200}{\hertz}$
    which we obtained empirically as part of~\cite{gangapurwala2020rloc}.
    
    As an example, $\pi^{8}_{b:2}$ represents a \textit{blind} motion control policy trained at \SI{8}{\hertz}.
    The state space of $\pi^{8}_{b:2}$ also contains $\mathbf{q}_j$ and $\mathbf{\dot{q}}_j$
    at joint recording steps $t_j-1$ and $t_j-2$, corresponding to a history length of 2.

    For brevity, we omit $f_t$ while referring to a class of motion control
    policies with the same operation mode and history length. For blind policies, $\pi_{b:0}$, with no joint state history,
    the state tuple $\mathbf{s}_{b:0}\in\mathbb{R}^{48}$ is defined as 
    
    \begin{equation*}
        \mathbf{s}_{b:0}:=\langle \mathbf{R}_B^T\mathbf{e}_z, \mathbf{q}_j, \mathbf{R}_B^T\mathbf{v}_{B}, \mathbf{R}_B^T\mathbf{\omega}_{B}, \mathbf{\dot{q}}_j, \Delta\mathbf{q}_j, \mathbf{c}^\ast \rangle,
    \label{eq:state_blind_history_0}
    \end{equation*}
    where $\mathbf{e}_z=[0, 0, 1]^T$ represents the vertical $z$-axis and $\mathbf{c}^\ast\in\mathbb{R}^{3}$ 
    comprises the desired heading velocity,
    lateral velocity and yaw rate commands represented in the base frame. 
    The objective of the motion control policies is thus to
    track user-generated desired velocity commands. 
    
    The state space of perceptive policies $\pi_{p:0}$ is written as $\mathbf{s}_{p:0}\in\mathbb{R}^{235}$.
    $\mathbf{s}_{p:0}$ augments $\mathbf{s}_{b:0}$ with robo-centric terrain information 
    $\mathbf{T}\in\mathbb{R}^{17\times{}11}$ observed between $[-0.8, 0.8]\,\si{\meter}$ along the heading
    axis and $[-0.5, 0.5]\,\si{\meter}$ along the lateral axis with a resolution of \SI{0.1}{\meter}. The 
    perceptive state space design is based on~\cite{rudin2022learning}.
    
    The joint state history augments the state space dimensionality by $H\times24$. For a blind policy with 
    history length of 4, $\pi_{b:4}$, its state
    tuple $\mathbf{s}_{b:4}\in\mathbb{R}^{144}$ is written as

    \begin{equation*}
        \mathbf{s}_{b:4}:=\langle \mathbf{s}_{b:0}, \mathbf{q}_{t_j-1}, \mathbf{q}_{t_j-2}, \mathbf{q}_{t_j-3}, \mathbf{q}_{t_j-4} \rangle.
    \label{eq:state_blind_history_4}
    \end{equation*}
    Here, $\mathbf{q}_{t_j}$ represents the joint state tuple comprising joint positions and joint velocities
    recorded at time step $t_j$. The control architecture,
    including the dense neural network policy architecture, is illustrated in Fig.~\ref{fig:control_architecture}.

\section{Methodology}
\label{sec:methodology}
\subsection{Training}
We represent our problem as a
sequential Markov decision process (MDP)~\cite{sutton1998introduction}
with the goal of obtaining a policy, or a class of policies,
that maximizes the expected cumulative discounted return,
\begin{equation}
    J\left(\pi\right)\doteq\underset{\mathcal{T}\sim\pi_\theta}{\text{\textup{E}}}\left[\sum_{t=0}^{N}{{\gamma}^{t}R}\right]\text{,}
\end{equation}
where $\gamma\in\left[0,1\right)$ represents the discount factor and $\mathcal{T}$, dependent on $\pi_\theta$,
denotes a finite-horizon trajectory with episode length $N$. Our reward function, 
$R$, comprises several reward
terms that allow for efficient and stable 
tracking of reference base velocity commands. We use the proximal
policy optimization (PPO)~\cite{schulman2017proximal} strategy to train
each of our policies. Our training approach, including the reward function, 
has been derived from prior 
works~\cite{hwangbo2019learning, gangapurwala2020guided, rudin2022learning}.

While our method is quite standard, training several policies for different
motion control frequencies requires tuning of individual
reward terms and hyperparameters such as $\gamma$. For example,
for an episodic length of \SI{1}{\second}, executing a policy
at \SI{200}{\hertz} would imply collection of forty times more
samples than for a \SI{5}{\hertz} policy. Additionally, the half-life of
$\gamma$ can be given by,
\begin{equation}
    n_{\gamma_{0.5}}=\frac{\log{}{0.5}}{\log{}\gamma}\approx \frac{-0.3}{\log{}\gamma}.
\end{equation}
For $\gamma=0.98$, the half-life would correspond to 34 control steps. For
a \SI{200}{\hertz} policy, this is equivalent to \SI{0.17}{\second} while
for a \SI{5}{\hertz} policy, this represents a duration of
\SI{6.8}{\second}.

To ensure consistency across different training frequencies, we denote the
duration of $n_{\gamma_{0.5}}$ in seconds as opposed to control steps. For a 
training frequency $f_t$, the discount factor can then be computed by
\begin{equation}
    \gamma = exp\left(\frac{\log{}{0.5}}{f_t \times n_{\gamma_{0.5}}}\right).
\end{equation}
In our training setup, we use $n_{\gamma_{0.5}}=\SI{3}{\second}$. For
an episodic length $N=\SI{1}{\second}$, we ensure the batch size per 
policy iteration remains the same for every control frequency. For this,
we perform parallel data collection wherein the number of parallel 
environments are scaled up to fit the desired batch size, $b_s=f_t\times{}n_{env}$.
For $b_s=\SI{48}{\kilo{}}$, we use $n_{env}=240$ for $f_t=\SI{200}{\hertz}$,
and $n_{env}=9600$ for $f_t=\SI{5}{\hertz}$.

To avoid retuning the reward function, we compute and aggregate the returns at each simulation step, $t_s$
as opposed to each control step $t_m$. Normally, $t_s\le t_m$ and we use $t_s=\SI{0.0025}{\second}$ in this
work. While this largely addresses
exhaustive reward function tuning, we observed that reward terms representing 
deviation from nominal joint configuration and action smoothness needed to be slightly tuned
for individual frequencies
to achieve visually similar locomotion behavior. We provide the different 
training configurations on the
\href{https://articulated.robots.ox.ac.uk/lfmc/}{project website}\footnote{\url{https://articulated.robots.ox.ac.uk/lfmc/}}.

\begin{figure}
        \centering
        \includegraphics[width=0.4\textwidth]{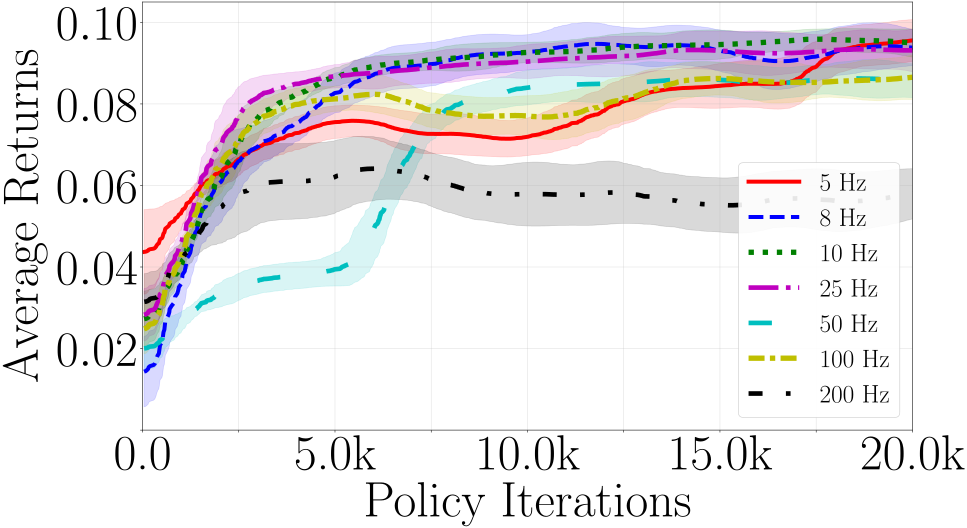}
        \caption{Average returns for each of the trained policies $\pi_{b:0}$.}
        \label{fig:average_returns}
        \vspace{-0.4cm}
    \end{figure}

We train each of the $\pi_{b:0}$ policies 
for \SI{20}{\kilo{}} iterations. The iteration
time is dependent on $f_t$ and varies between
\SI{0.4}{\second} (for $f_t=\SI{25}{\hertz}$ and $f_t=\SI{50}{\hertz}$) to
\SI{1.5}{\second} (for $f_t=\SI{5}{\hertz}$ and $f_t=\SI{200}{\hertz}$).
on a standard desktop computer housing an 8-core \SI{3.6}{\giga\hertz} Intel
i9-9900K and an Nvidia RTX 2080Ti. The returns plot for
each of the trained policies is shown in Fig.~\ref{fig:average_returns}. 
The policies trained at low-frequencies (\SI{8}{\hertz}, \SI{10}{\hertz} and \SI{25}{\hertz})
converge a lot faster ($<$10k iterations) compared to high-frequency policies. 
Note, $\pi^{5}_{b:0}$ suffers from poor reactivity 
and is therefore harder to train.

We do not perform any dynamics randomization (DR) while training the blind policies.
Although we do use an actuator network~\cite{hwangbo2019learning} to model the 
real actuation dynamics, in Section~\ref{sec:results-and-discussion}, we
show that introducing the actuator network during training
may not even be necessary for LFMC.

\subsection{Evaluation}
We follow the narrative of bio-inspired low-frequency motion control (LFMC)
and discuss the following key observations and reasoning in Section~\ref{sec:results-and-discussion}.
\begin{itemize}
    \item LFMC policies are less sensitive to actuation dynamics under the assumption that
    actuation settling time~\cite{vukosavic2012electrical} is less than control step time (Section~\ref{sec:results:intuitive_reasoning}).
    \item LFMC policies do not perform implicit modeling of system dynamics necessary for predictive control at high frequencies.
    Instead, LFMC policies can operate as motion planners (Section~\ref{sec:results:behavioral}). To support this, we visualize the policy
    network Jacobians in Fig.~\ref{fig:jacobian_means}.
    \item Since LFMC policies operate as motion planners, they show more robustness to variations in system dynamics. This is based
    on the assumption that the low-level actuation tracker stably and reliably tracks the motion plans. We show this to be the case in Fig.~\ref{fig:success_rate_corresponding}.
    \item LFMC policies are faster to train (Fig.~\ref{fig:average_returns}).
\end{itemize}

\begin{figure}
        \centering
        \includegraphics[width=0.48\textwidth]{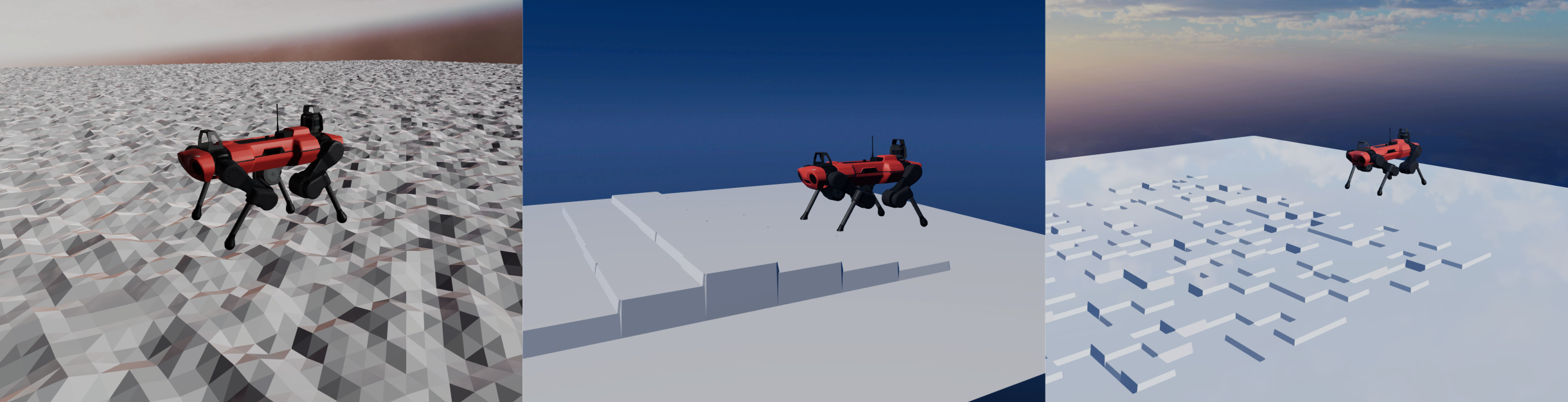}
        \caption{RaiSim simulation set up for evaluation of blind and perceptive locomotion policies 
        with ANYmal C traversing terrains comprising unstructured ground, stairs and bricks.}
        \label{fig:raisim_experiments}
        \vspace{-0.4cm}
    \end{figure}

To support these points, we evaluate the performance of each of the 
individual blind $\pi_{b:0}$ policies in RaiSim with unstructured rough terrain 
generated using Perlin noise~\cite{perlin2002improving} with maximum extrusion
of \SI{0.15}{\meter}. This is shown in Fig.~\ref{fig:raisim_experiments}.
Our motivation for this setup is twofold: (1) the terrain noise
introduces randomness allowing us to measure a probability distribution and
(2) the unexpected perturbations highlight the reactivity of each of the policies.

We introduce success rate (SR) as a performance metric defined as,
\begin{equation}
    \mathrm{SR} = 1 - \frac{N_e}{N_T}
\end{equation}
where $N_e$ refers to the number of episodic rollouts that were 
terminated early due to an invalid robot state and $N_T$ represents
the total number of rollouts. In this work, we use $N_T=100$. 
For each rollout, we randomize the base heading direction. This randomization
occurs with the same seed across each of the policies. An invalid
robot state is defined by the criteria: (1) $\arccos{(\mathbf{R}_{B_{3,3}})} > 0.4\pi$ which
relates to base orientation, 
(2) self-collisions, or (3) collision of the robot base with ground.

We train and compare $\pi^{10}_{b:4}$ and $\pi^{200}_{b:4}$ to show that joint state history
is relevant for modeling system dynamics and is essential for high-frequency motion control
as presented in~\cite{hwangbo2019learning}. This, however, is not the case for LFMC. We also evaluate the performance of perceptive locomotion policies
on terrains comprising rough ground, stairs and bricks as shown
in Fig.~\ref{fig:raisim_experiments}. Our evaluation method for perceptive locomotion policies
is based on the
setup introduced in~\cite{gangapurwala2020rloc}.

\section{Results}
\label{sec:results-and-discussion}
This section presents the key results in support
of our contribution. We provide 
an extended analysis on the \href{https://articulated.robots.ox.ac.uk/lfmc/}{project 
website}. The \href{https://articulated.robots.ox.ac.uk/lfmc/}{project 
website} also contains qualitative demonstrations of \SI{8}{\hertz} motion 
control in non-stationary environments such as unexpected slippery 
surfaces.

\subsection{Intuitive Reasoning}
    \label{sec:results:intuitive_reasoning}
\begin{figure}
        \centering
        \includegraphics[width=0.45\textwidth]{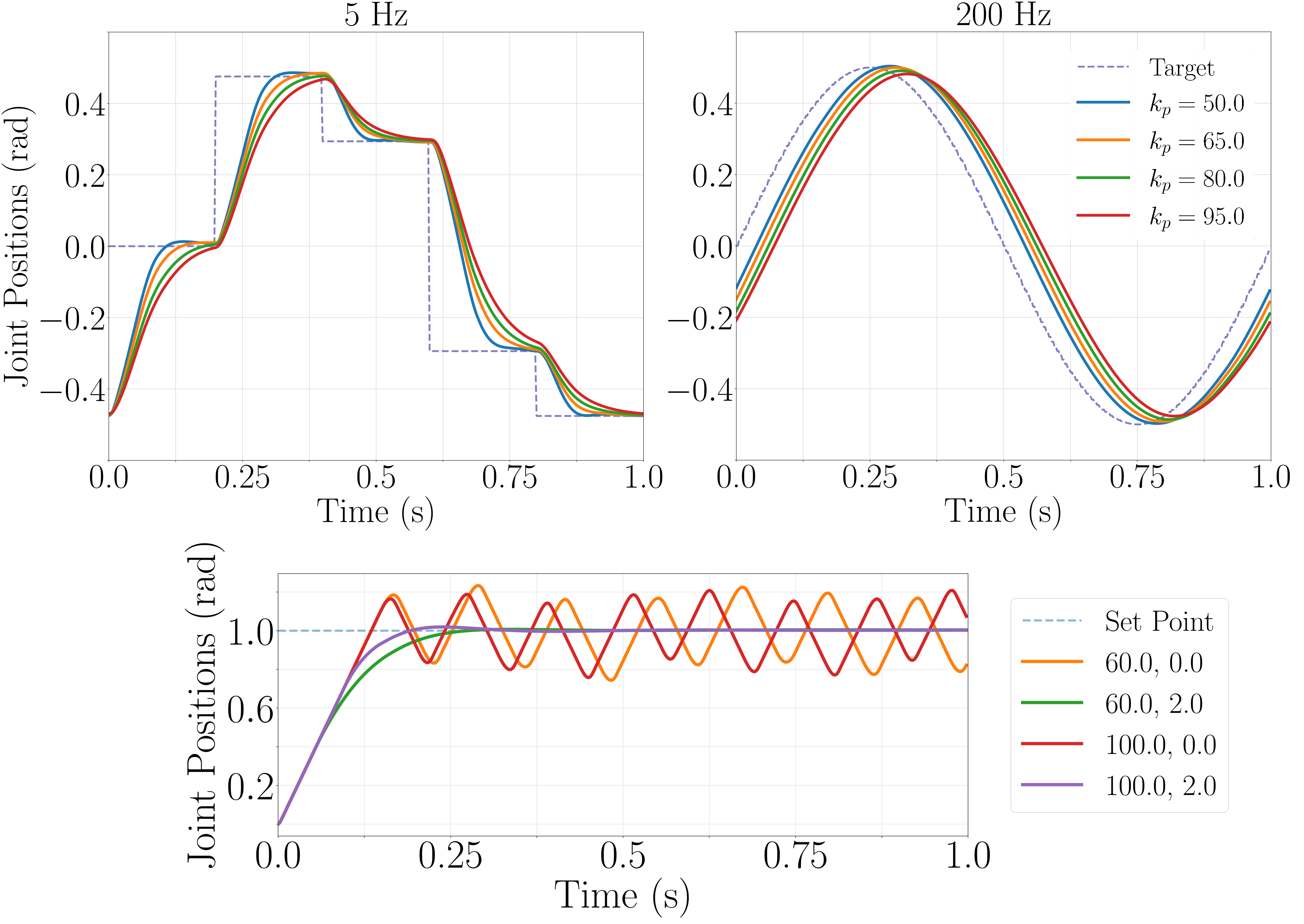}
        \caption{\textit{Top}: Tracking of sinusoidal set points updated at \SI{5}{\hertz} and \SI{200}{\hertz}
        for various position tracking gains. \textit{Bottom}: Step responses observed for $k_p\in\{60, 100\}$ and $k_d\in\{0, 2\}$
        for a series elastic actuator present on the ANYmal C quadruped.}
        \label{fig:joint_trajectories_vs_frequencies}
    \end{figure}

    Figure~\ref{fig:joint_trajectories_vs_frequencies} (top) illustrates a toy example
    of a 1 DoF PD controller tracking 
    sinusoidal set points updated at \SI{5}{\hertz} and \SI{200}{\hertz} 
    for $k_p\in\{50, 65, 80, 95\}$ and $k_d=2$.
    For the \SI{5}{\hertz} update frequency, the joint trajectories 
    converge to very similar states before a new set point is generated. Note that, we
    use $k_p=80$ and $k_d=2.0$ for deploying our policies on to the real robot. For a 
    \SI{5}{\hertz} controller, this implies the sensory readings at each update step
    are less effected by actuation tracking dynamics in comparison
    for higher update frequencies such as \SI{200}{\hertz}.
    This implies that, for an effective control behavior, the \SI{200}{\hertz}
    policy requires observability of the actuation dynamics.
    
    We hypothesize that LFMC allows for operation as a planner and 
    refer to it as \textit{motion planning hypothesis}. In this context,
    motion planning refers to generation of target states without an 
    adaptive tracking system as is common in optimization-based 
    approaches~\cite{Bellicoso2018} which utilize a planner in addition to 
    a whole-body controller. This makes low-frequency
    motion controllers robust to actuation dynamics under the assumption 
    that the 
    low-level actuation controller stably tracks the generated joint states. 
    Figure~\ref{fig:joint_trajectories_vs_frequencies} (bottom) illustrates why the stable tracking
    is necessary. For cases of under or over-damping, the motion controller may be required to adapt to
    the settling state even with low-frequency control.
 
     \begin{figure}
        \centering
        \includegraphics[width=0.39\textwidth]{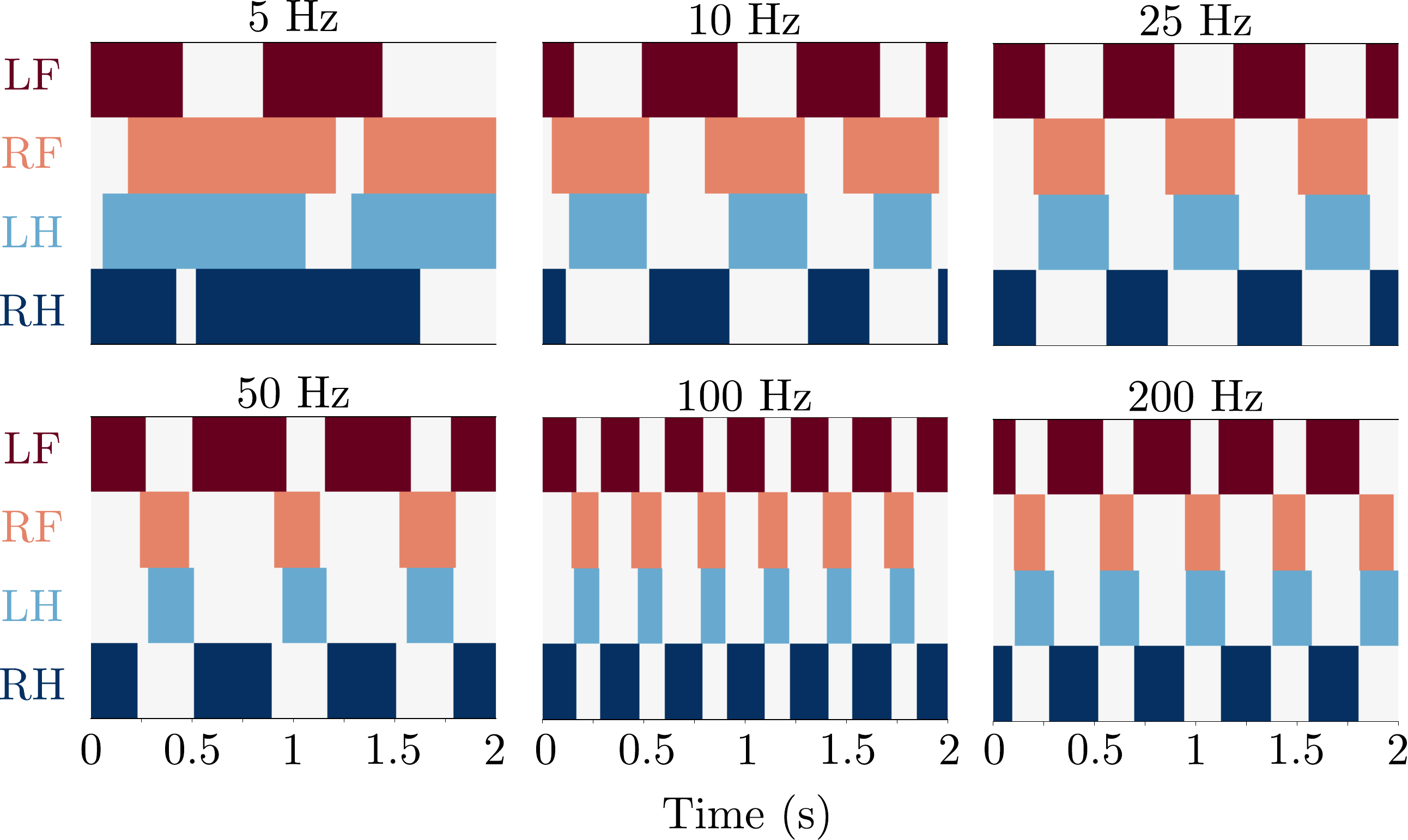}
        \caption{Gait sequences for various $\pi_{b:0}$ motion control policies. The coloured
        regions represent stance phase. }
        \label{fig:contact_sequences}
        \vspace{-0.4cm}
    \end{figure}

    \begin{figure}
        \centering
        \includegraphics[width=0.40\textwidth]{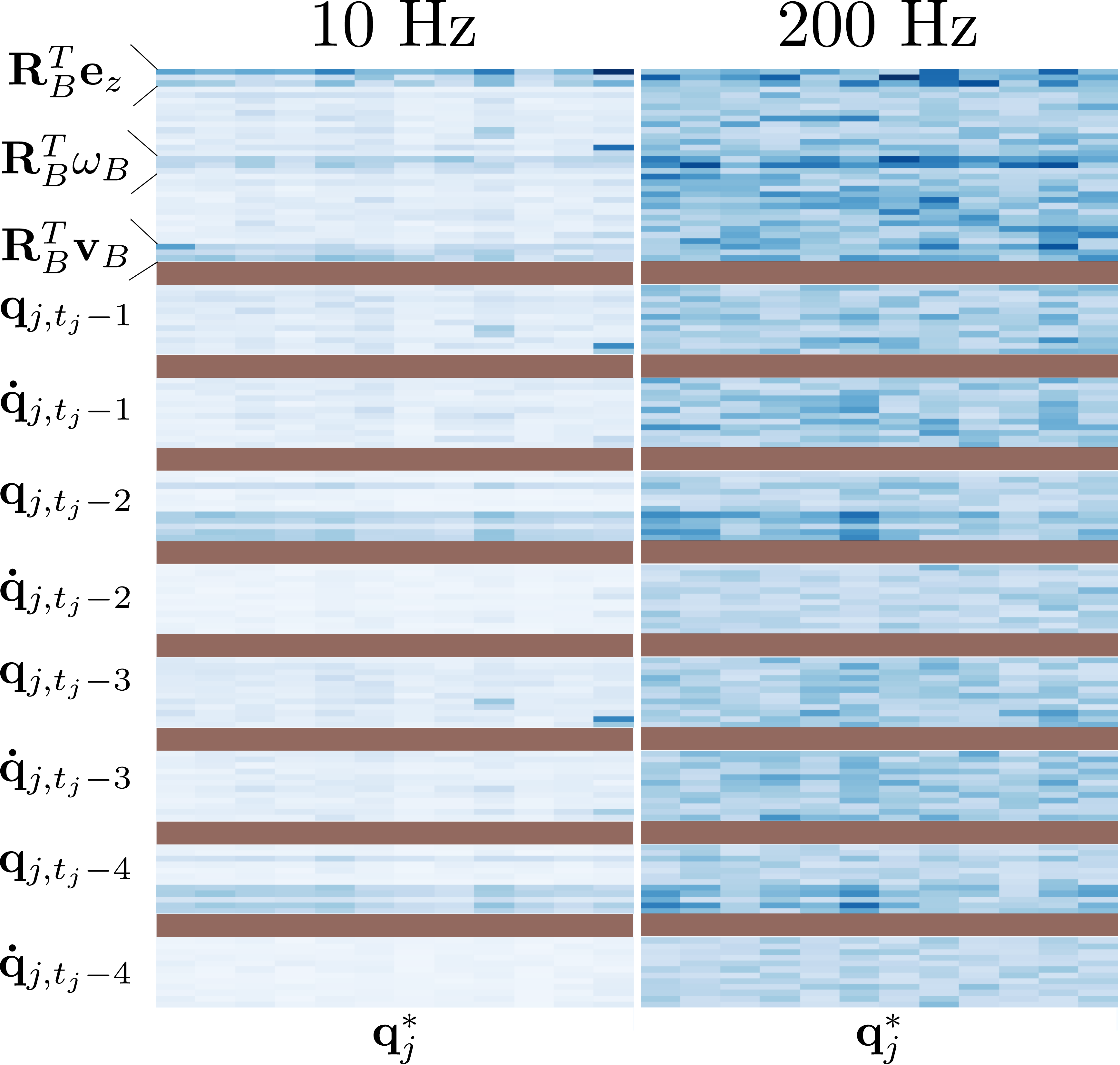}
        \caption{Visualization of the mean of network Jacobians recorded for $\pi^{10}_{b:4}$ and $\pi^{200}_{b:4}$ for \SI{2}{\second}.
        Dark blue regions correspond to high gradients whereas white corresponds to zero gradients. The brown regions separate 
        different observations and have only been included for visual aid. Note that, joint state history is 
        sampled at \SI{200}{\hertz} for both the policies. Sampling joint state history at \SI{10}{\hertz} for $\pi^{10}_{b:4}$
        resulted in near-zero gradients for history terms.}
        \label{fig:jacobian_means}
        \vspace{-0.5cm}
    \end{figure}

    \begin{figure*}
        \centering
        \includegraphics[width=0.92\textwidth]{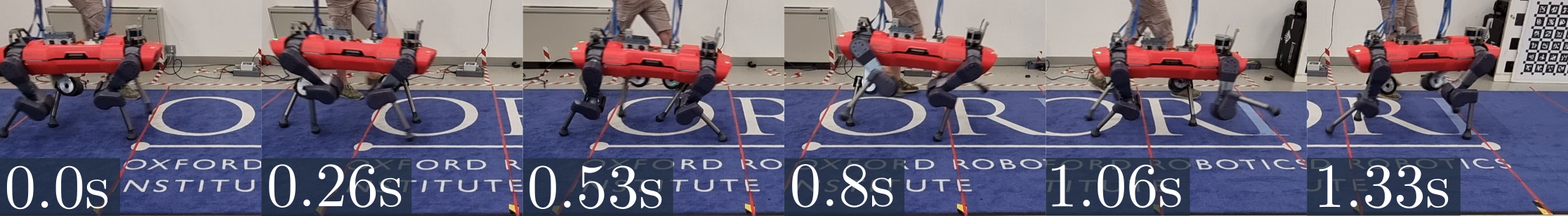}
        \caption{Motion control policy trained and deployed at \SI{8}{\hertz} stably tracking heading base 
        velocity of \SI{1.5}{\meter\per\second}.}
        \label{fig:high_speed_locomotion_frames}
        \vspace{-0.2cm}
    \end{figure*}

    \begin{figure}
        \centering
        \includegraphics[width=0.45\textwidth]{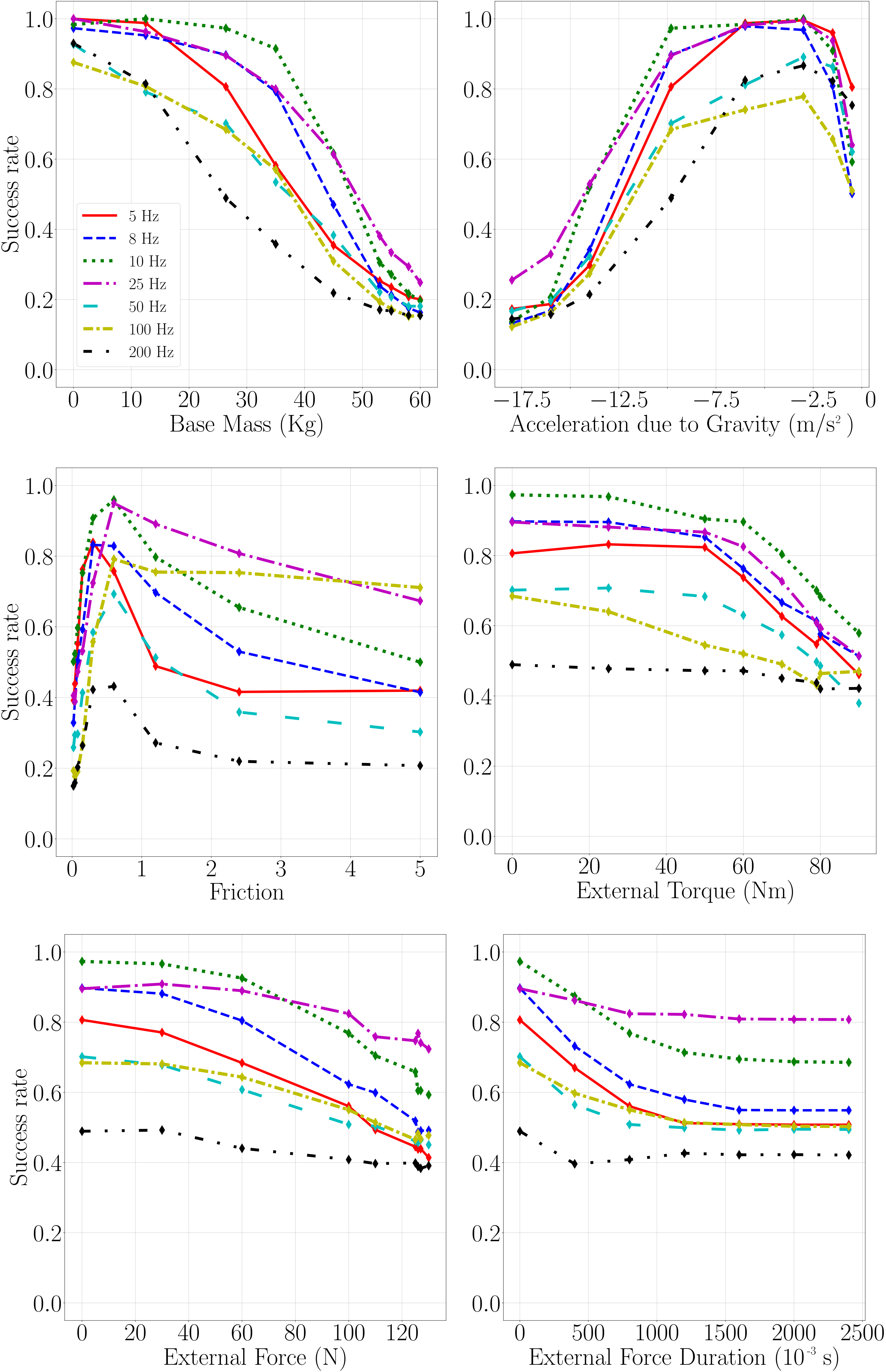}
        \caption{Success rate observed for various motion control policies, $\pi_{b:0}$, for 
        different perturbations and dynamics parameters.}
        \label{fig:success_rate_corresponding}
        \vspace{-0.5cm}
    \end{figure}

\subsection{Qualitative and Behavioral Analysis}
~\label{sec:results:behavioral}
    We test the motion planning hypothesis by transferring
    the trained motion control policies on to the 
    real ANYmal C quadruped. We observe extremely aggressive actuation
    tracking for $\pi^{200}_{b:0}$ resulting in vibrations
    at the rotary joints. This aggressive 
    behavior is reduced with lower-frequency policies and no vibrations are recorded
    for $\pi^{25}_{b:0}$ and lower. 
    We suspect that, since no dynamics randomization (DR) was performed during training, and
    due to imperfect actuation modeling, high-frequency policies overfit to the
    simulation domain affecting sim-to-real transfer. Note, 
    while we are able to transfer $\pi^{5}_{b:0}$
    onto the real robot, we
    are only able to stably execute low-velocity motions. The \SI{5}{\hertz}
    policy suffers from poor reactivity and is unable to execute recovery
    actions in unstable states.

    We observe interesting behavior with regards to the stance (foot-in-contact) and 
    swing (foot-not-in-contact) phase duration.
    Low-frequency policies exhibit larger stance and swing phases
    compared to high-frequency policies (Fig.~\ref{fig:contact_sequences}). 
    We expected this to be an artifact of the scaling of action smoothness reward term
    (which penalizes large deviations between current and previous actions) with variations
    in motion control training frequencies. This,
    however, was not the case when we introduced joint state history ($N\ge4$) into the state space.
    \textit{Hwangbo et al.} hypothesized that the joint state history implicitly modeled contact
    detection~\cite{hwangbo2019learning}. While this has been consistent with our
    analysis of observing the absolute of policy Jacobians, $\abs{d\pi_{\theta}(s)/d{}s}$, as
    presented in~\cite{Leeeabc5986}, we also observed that high-frequency
    control policies are more dependent on joint state history than low-frequency policies. 
    We posit that joint state history
    improves the domain observability through implicit encoding of actuation dynamics~\cite{akkaya2019solving}
    and is therefore more relevant for high-frequency policies.
    
    We further investigate this for $\pi^{10}_{b:4}$ and $\pi^{200}_{b:4}$. 
    Figure~\ref{fig:jacobian_means} illustrates the mean of the policy Jacobians recorded
    for a duration of \SI{2}{\second}.
    Dark blue regions suggest higher gradients,
    implying larger dependency. Compared to \SI{10}{\hertz}, the \SI{200}{\hertz} policy
    requires more observations to execute the same task relating to larger dependency
    on system dynamics. Interestingly, the gradients for joint velocities are
    quite low for $\pi^{10}_{b:4}$ while $\pi^{200}_{b:4}$ utilizes joint velocities
    more than joint positions. The gradients observed along the base 
    velocity states were also negligible for \SI{10}{\hertz} policy, 
    compared to \SI{200}{\hertz}, even during high-speed locomotion 
    suggesting that LFMC policies do not considerably rely on base-velocity measurements.
    
    For high-frequency $\pi_{b:0}$, we suspect the fast contact switching behavior occurs due to partial 
    observability of the system dynamics. In our experience, we have observed this to be the
    case for poorly designed state spaces. This, however, needs further investigating. 
    For the \SI{200}{\hertz} policy, the increase in
    stance and swing phase duration, compared to \SI{100}{\hertz} policy, is due to poor tracking with lower stability. 

\subsection{Robustness Analysis}
    One of our main objectives with this work is to demonstrate
    robustness with low-frequency motion control. 
    Figure~\ref{fig:success_rate_corresponding} shows that
    $\pi^{10}_{b:0}$ performs better than most
    of the policies. $\pi^{25}_{b:0}$ offers both high robustness
    and reactivity whereas
    $\pi^{8}_{b:0}$ falls behind $\pi^{10}_{b:0}$ due to
    inadequate reactivity for traversal
    over rough terrain.
    We also investigate robustness to actuation latencies and show that
    LFMC policies offer higher robustness than high-frequency policies (Table~\ref{table:actuation_latencies}).

        \begin{center}
        \begin{table}[htbp]
            \caption{Maximum actuation delay that $\pi_{b:0}$ policies can be robust to right before failure measured
            at a resolution of \SI{5}{\milli\second}.}
            \resizebox{0.48\textwidth}{!}{
            \centering
            \begin{tabular}{ |c|c|c|c|c|c|c| }
                \hline
                \textbf{Training Frequency (Hz)} & \textbf{5} & \textbf{10} & \textbf{25} & \textbf{50} & \textbf{100} & \textbf{200} \\
                \hline
                \textbf{Latency (ms)} & 90 & 90 & 65 & 50 & 30 & 20 \\
                \hline
            \end{tabular}}
            \label{table:actuation_latencies}
        \vspace{-0.3cm}
        \end{table}
        \end{center}

\subsection{Dynamic Locomotion}
    We perform qualitative evaluation on the real robot
    and demonstrate high-speed dynamic locomotion with $\pi^{8}_{b:0}$.
    As shown in Fig.~\ref{fig:high_speed_locomotion_frames}, 
    we are able to achieve a heading velocity of approximately
    \SI{1.5}{\meter\per\second} traversing a distance of \SI{2}{\meter} in roughly
    \SI{1.33}{\second}.
    
    We also train a perceptive locomotion policy $\pi^{8}_{p:0}$ based on~\cite{rudin2022learning}. We show that \SI{8}{\hertz} motion
    control is sufficient for robust traversal over considerable obstacles (wooden railway sleepers) and steps (both up and down) as presented in Fig.~\ref{fig:anymal_c_lab_terrain}.
    
    We further compare the behavior of policies trained with and without 
    DR. The DR parameters are based on~\cite{gangapurwala2020guided} and have been provided on
    the \href{https://articulated.robots.ox.ac.uk/lfmc/}{project website}.
    This is shown for \SI{10}{\hertz} and \SI{200}{\hertz} perceptive policies in Table~\ref{table:perceptive_success_rate}.
    As expected, DR allows for better performance over uneven terrain. Introduction of joint state history is
    not as effective as doing both, introducing joint state history and DR. Joint state history and DR allow
    for better observability of environment interactions, necessary for uneven terrains~\cite{Leeeabc5986}, while also encouraging 
    generalizability to unseen domains.
    
    We are also able to demonstrate transfer on to the physical system with policies trained without the
    actuator network. The behavior is stable, however, not as smooth as policies trained with the 
    actuator network. We detail upon the extended analysis on the \href{https://articulated.robots.ox.ac.uk/lfmc/}{project website}
    and summarize our evaluation in the overview video.

        \begin{center}
        \begin{table}[htbp]
            \caption{Success rates of \SI{10}{\hertz} and \SI{200}{\hertz} 
            perceptive policies measured for 100 runs each.}
            \resizebox{0.48\textwidth}{!}{
            \centering
            \begin{tabular}{ |c|c|c|c|c||c|c|c|c| }
                \hline
                 & \multicolumn{4}{c||}{\textbf{\SI{10}{\hertz}}} & \multicolumn{4}{c|}{\textbf{\SI{200}{\hertz}}} \\
                 & $\pi_{p:0}$ & $\pi_{p:0}$ (DR) & $\pi_{p:4}$ & $\pi_{p:4}$ (DR) & $\pi_{p:0}$ & $\pi_{p:0}$ (DR) & $\pi_{p:4}$ & $\pi_{p:4}$ (DR) \\
                \hline
                \textbf{Rough} & 0.94 & 0.94 & 0.94 & 0.95 & 0.87 & 0.92 & 0.93 & 0.94 \\ 
                \hline
                \textbf{Stairs} & 0.86 & 0.93 & 0.92 & 0.94 & 0.52 & 0.59 & 0.88 & 0.95 \\ 
                \hline
                \textbf{Bricks} & 0.66 & 0.71 & 0.69 & 0.80 & 0.58 & 0.62 & 0.64 & 0.76 \\ 
                \hline
            \end{tabular}}
            \label{table:perceptive_success_rate}
        \vspace{-0.3cm}
        \end{table}
        \end{center}

\section{Conclusion}
\label{sec:conclusions-and-future-work}
This work aims to start the discussion within the robotics community about the role and benefit of high- versus low-frequency motion control, especially within the context of learning-based approaches. From biological studies we know that animals can perform robust and dynamic locomotion at low motion control frequencies and, with this work, we showed how real robots can achieve this too. 

We demonstrated dynamic and robust quadrupedal locomotion with as
low as \SI{8}{\hertz} of motion control frequency. We further provided
empirical evaluations to support our claim that motion control
policies trained at low-frequencies do not require dynamics randomization
or actuation modeling to perform a successful sim-to-real transfer.


\clearpage

\bibliographystyle{IEEEtran}
\bibliography{references.bib}

\end{document}